\documentclass[preprint, 12pt]{elsarticle}
\usepackage{amsmath}
\usepackage{amssymb}
 \usepackage{amsthm}
 \usepackage{hyperref}
 \usepackage{subfigure}
 \usepackage{algorithmic}
 \usepackage{algorithm}
\usepackage{booktabs}
    \makeatletter
    \def\ps@pprintTitle{%
    	\let\@oddhead\@empty
    	\let\@evenhead\@empty
    	\let\@oddfoot\@empty
    	\let\@evenfoot\@oddfoot
    }
    \makeatother
\newtheorem{thm}{Theorem}

\newdefinition{rmk}{Remark}
\newproof{pf}{Proof}
\newdefinition{defi}{Definition}
\begin{document}
	
\begin{frontmatter}
%
\title{Relevant based structure learning for feature selection}

 \author[rv1]{Hadi Zare \corref{cor1}}
\author[rv1]{Mojtaba Niazi}
\cortext[cor1]{Corresponding author}
\address[rv1]{Faculty of New Sciences and Technologies, University of Tehran, Iran}

\begin{abstract}
Feature selection is an important task in many problems occurring in pattern recognition, bioinformatics, machine learning and data mining applications. The feature selection approach enables us to reduce the computation burden and the falling accuracy effect of dealing with huge number of features in typical learning problems.
There is a variety of techniques for feature selection in supervised learning problems based on different selection metrics.
In this paper, we propose a novel unified framework for feature selection built on the graphical models and information theoretic tools. The proposed approach exploits the structure learning among features to select more relevant and less redundant features to the predictive modeling problem according to  a primary novel likelihood  based criterion.  In line with the selection of the optimal subset of features through the proposed method, it provides us the Bayesian network classifier without the additional cost of  model training on the selected subset of features.  The optimal properties of our method are established through empirical studies and computational complexity analysis. Furthermore the proposed approach is evaluated on a bunch of benchmark datasets based on the well-known classification algorithms. Extensive experiments confirm the significant improvement of the proposed approach compared to the earlier works.
\end{abstract}


\end{frontmatter}


\section{Introduction}
\label{intro}
Feature selection (or variable selection) has been considered as a primary step in machine learning, pattern recognition, and data mining fields. It is used in a variety of applied domains such as text classification, micro-array analysis and image processing. 
Nowadays with the explosion of massive online data, choosing an optimal subset of features is a very crucial step, \cite{wang_onlineFS_2014, tan_towardsUltraHigh_2014}. 
While predictive modeling with huge feature sets are  common  in recent years, it would be caused  heavy computational burden, interpretation difficulty, and weak results based on curse of dimensionality \cite{bishop_pattern_2007, ida13}. 
Not only the suitable feature selection process can provide efficient tools to remove  irrelevant, redundant and noisy features, but it would improve the speed of learning phase and performance measures of the predictive task too.
Based on learning language, the feature selection could be classified to supervised and unsupervised methods.
Supervised feature selection approaches are mainly based on the relation between the features and the label to find the optimal feature sets \cite{Moradi_relevanceredundancy_2015} \cite{peng_MRmr_2005} \cite{wu_osfs_2013}\cite{zokaei_ashtiani_bandit-based_2014} \cite{wang_feature_2015}. 
On the other hand, finding the optimal feature selection techniques for unsupervised problems are much harder than the supervised one's due to the ambiguous definition of the unsupervised learning, ``discovering the interesting patterns from the data'' \cite{murphy_machine_2012} \cite{feng_unsupervised_2016} \cite{moradi_graphUnsFS_2015}\cite{tabakhi_unsupervised_2014}\cite{perkins_online_2003}. 
Here we concentrate on the feature selection for the supervised learning problems. \par

If we let the original feature set $F = \{f_1, f_2, \dots, f_p \}$ and the class variable as $Y$, the aim of feature selection process is to find the optimal  subset $S\subset F$ such that it has the best predictive accuracy based on the validation performance criteria.
Supervised feature selection process typically can be divided in four primary steps \cite{liu_toward_2005}, 
\begin{itemize}
	\item[(i)] Evaluation criteria
	\item[(ii)] Search approaches 
	\item[(iii)] Stopping criterion 
	\item[(iv)] Validation methods
\end{itemize}
 In evaluation step, a criterion should be designed carefully to test the relevancy between the selected subset of features and  the class variable. Because of the exponential computational complexity of searching through the complete subsets of the original set of features, search procedure for generating candidate subsets of features to evaluate them are devised in search step.  The search and evaluation on the candidate subset of features are continued until the stopping criterion holds. Finally the selected feature set usually requires to be validated based on the dataset or prior domain expert knowledge. The evaluation criteria and search strategy are more important than the other steps in a feature selection process. \par
 Based on different evaluation criteria, the feature selection techniques can be generally classified  into three main types, the filter, the wrapper and the hybrid methods, \cite{guyon_introduction_2003} \cite{liu_toward_2005}\cite{DashLiu97-ida}\cite{liu_feature_2015}.
The straightforward  approach for evaluation criteria is to  measure directly the performance of  a subset of features based on classification accuracy with the aid of  a predictive classifier to select the best subset of features \cite{KohaviJ97-ai}. Although the most effective and optimal approach could be offered in a wrapper model, these techniques suffer from heavy computational burden of training classifier algorithms. 
The main idea of filter methods is the selection of the optimal features based on statistical or information theoretic  evaluation criteria applied on the certain characteristics of the data without requirement of any classification algorithms.
The hybrid (embedded) techniques that are somewhat similar  but less computationally expensive compared to wrapper methods which  measure optimal subset of features through the learning phase. 
Because of the time consuming of the wrappers and hybrid techniques, the filter methods are highly recommended for dealing with real applications using a variety of evaluation tools such as,  the Markov blanket based for streaming dataset \cite{wu_osfs_2013}, fuzzy-rough sets for feature significance \cite{MajiG13-tcyb},  heuristic relevance based approach \cite{moradi_integration_2015},  divergence criterion \cite{BressanV03-pami}, and centrality based influence measure  \cite{moradi_graphUnsFS_2015}.\par

A variety of techniques are proposed for search strategies, such as exhaustive search \cite{OliveiraS92-icml}, ranking based among the feature based on the relevancy to the class variable \cite{Lewis92} \cite{Geng07--FSRanking} \cite{PinheiroCCR12-eswa}. 
Because of the exponential computation time of exhaustive search approach and ignoring the redundant features in relevant ranking based methods,  sequential greedy approaches are proposed  to maximize the  evaluation criteria  in an iterative and incremental development manner \cite{Liu1998-Book}\cite{guyon_introduction_2003}.   Although  the traditional forward greedy approaches are commonly used for dealing with huge number of features because of low computational burden and more robustness to over-fitting, they suffer from neglecting the impact of redundancy among features. Some methods  \cite{koller_toward_1995} \cite{peng_MRmr_2005} have proposed the innovative information theoretic evaluation criteria in a sequential search vein to remedy the aforementioned problems. \par
Recently some feature selection methods are proposed for massive online dataset, where the number of features are increased with fixed number of observations, streaming cases or incremental observations \cite{wu_osfs_2013}\cite{wang_onlineFS_2014}\cite{wu_massive-scale_2014} and in these works the evaluation criteria is based on the priorly defined probabilistic and information theoretic concepts  in  \cite{koller_toward_1995, peng_MRmr_2005}.
The main problems in these recent works could be categorized in threefold, computational burden, streaming setting and optimality criteria.\par
In this paper we propose a novel feature evaluation criteria in a filter approach based on structure learning and information theoretic tools that can be adopted for streaming dataset and non-streaming dataset. 
In line with the proposed approach, \emph{``structure learning for feature selection''}, hereafter called as \emph{SLFS}, that allowed us to choose more relevant less redundant features  carefully within a negligible loss of total feature information,  the computation time is reduced compared to the earlier works such as  \cite{peng_MRmr_2005, wu_osfs_2013}. \par
The structure of the paper is organized as follows. In Section \ref{Sec2}, the related works on feature selection are reviewed and a motivation of the basic idea to solve the problem are presented. Section \ref{Sec3}  is devoted to the theoretical foundation of the feature selection based on the Markov blanket approach and an overall scheme of our method. The proposed feature selection algorithms and their advantages compared to the previous ones are illustrated in Section \ref{Sec4}. We present and describe the experimental results based on the state-of-the art datasets through the SLFS algorithm compared to the earlier works in Section \ref{Sec5}. Finally Section \ref{Sec6} discusses the results based on the proposed framework as well as conclusions and future works on the field.
\section{Related works and Basic Idea}
\label{Sec2}
\subsection{Related works}
Because of the importance of the feature selection problem, many researches have been done on various aspects of this fundamental topic. By the availability of massive number of features,   reasonable to assume  a large subset of  features are either irrelevant or redundant for predictive modeling and only a small portion of relevant features yield more effective learning aims \cite{DashLiu97-ida}\cite{liu_toward_2005}. On the one hand, most of the earlier researches  have been concentrated on finding relevant features based on the high dependency to the class labels \cite{peng_MRmr_2005}\cite{BressanV03-pami}\cite{HuPYL10-ieee-tsmc}. On the other hand, for a wide variety of applications, such as genomic microarray analysis \cite{XingJK01-icml} \cite{baur_feature_2016}, image representation \cite{MichaelElad08-SIAM},  and text categorization \cite{Forman2003-jmlr}\cite{hava_supervised_2013}, there exist high redundancy among the features. Hence the feature selection algorithm based only on the relevance criteria can be resulted in suboptimal set of features \cite{KohaviJ97-ai}\cite{PinheiroCCR12-eswa}. There are many research efforts to consider the feature selection criteria with the redundancy and relevancy simultaneously, Markov blanket based approach \cite{koller_toward_1995}\cite{liu_information-theoretic_2015},  max-dependency and min redundancy based on mutual information \cite{peng_MRmr_2005}, and meta-heuristic greedy search \cite{Moradi_relevanceredundancy_2015}.     
From the theoretical point of view,  Markov blanket framework for feature selection would  be yielded to the optimum subset of features and the remaining ones could be considered as redundant features. Because of the exponential computational complexity of  finding  the Markov blanket subset among the features, there exist a variety of efforts to approximate it such as linear correlation  approach \cite{Liu04-jmlr},  and statistical  $\chi^2$-square test \cite{wu_osfs_2013}. Those works consider pairwise  feature dependency rather than the joint consideration to find the Markov blanket. 
\subsection{Overall scheme of the idea}
First, we define the mutual information between two features $x$, $y$, 
\begin{align}
\label{eq1}
I(x; y)&=\int_{x}\int_{y}\,P( x, y)\log \frac{P( x, y)}{P( x)P( y)}dx~dy\cr
&= H(x)-H(x|y)
\end{align}
where  $H(x)$ is the entropy of the $x$ and $H(x|y)$ is the conditional entropy of $x$ given $y$, \cite{vergara_review_2013}.
The definition \eqref{eq1} of mutual information can be generalized for a vector of features $E = (f_1, f_2, \dots, f_m)$ and class variable $Y$ as follows,
\begin{equation}
I( E; Y)=\iint_{E}\!\!\int_{Y}\!\!\! P( f_1,\dots, f_m, Y)\log\!\frac{P( f_1, \dots, f_m, Y)}{P( f_1)\dots P(f_m)P( Y)}df_1\!\cdots df_m dY
\end{equation}
 By denoting the total feature set  $F = \{f_1, f_2, \dots, f_p \}$  and a selected subset of it  as $S\subset F $, the optimal solution for a supervised feature selection is to reduce the size of selected features $S$ such that it produces the most prediction accuracy with the class variable \cite{vergara_review_2013},
 \begin{equation}
 \label{eq2}
 R = \min_{S\subset F} |S|-\lambda . I(S;Y)
 \end{equation}
 where  $\lambda>0$ is a penalized constant. The main problem with this approach is twofold, the computation of the joint mutual information  and the problem of remaining redundant when new features are added in each step via a greedy approach like \cite{peng_MRmr_2005}. Unlike to the stepwise selection and removal of the features, in line with the subset based strategy to investigate the optimal features,  we propose a model based technique to find the optimum subset of features. The main idea of  proposed method  is to find the structure of directed graphical models among the features  to provide a suitable framework for extraction of the selected features  based on the well-known Markov blanket vehicle. 
 We have applied maximum likelihood approach  for structure learning among the features that yields to a unique solution. 
 The details of our approach are presented in the following sections.
\section{Theoretical foundation for the proposed approach}
\label{Sec3}
Because of our proposed method's dependency on Markov blanket and conditional independency concepts, the precise definitions are 
presented here based on \cite{Koller_probabilistic_2009}.
\begin{defi} \label{def1}
(conditional independence) Let $X,\, Y,\, Z$ be sets of discrete random variables.  $X$ is conditionally independent of $Y$ given $Z$, $X\perp Y|Z$, if $$P(X=x|Y=y, Z=z) = P(X=x|Z=z) \quad \textrm{or} \quad P(Y=y, Z=z)=0 $$  
\end{defi}
If we denote the $f_i\in F$ and $S_i = F-\{f_i\}$, then the formal definitions of relevant and redundant features are as follows \cite{Liu04-jmlr}.
\begin{defi} \label{def2_StRel}
 	(Strongly relevant) A feature $f_i$ is said to be strongly relevant to the class variable $Y$ if and only if 
 	$$\nexists S_i \quad\textrm{such that}\quad P(Y|f_i, S_i)= P(Y|S_i)$$
\end{defi}
\begin{defi} \label{def3_WeRel}
  	(Weakly relevant) A feature $f_i$ is said to be weakly relevant to the class variable $Y$ if and only if 
  	$$\exists S_i \quad \textrm{such that}\quad P(Y|f_i, S_i) =  P(Y|S_i)$$
\end{defi}
\begin{defi} \label{def3_Ire}
	(Irrelevant) A feature $f_i$ is said to be irrelevant to the class variable $Y$ if and only if 
	$$\forall S_i \quad\textrm{such that}\quad P(Y|f_i, S_i) =  P(Y|S_i)$$
\end{defi}
As stated in definitions, while strongly relevant features provide unique information about the class variable which are not attainable with the other features, weakly relevant features have information about the class variable  attainable with the other features without losing probabilistic information. On the other hand,  irrelevant features are not related to the class variable $Y$ and they should be removed from the modeling.  Although the definition \ref{def3_WeRel} states weakly relevant features maybe useful to the class variable and  therefore  including some of them in the optimal subset of features, the discrimination between these features is not clear based on the relevancy concept. In  works of \cite{koller_toward_1995} and  \cite{Liu04-jmlr} the concept of redundant features based on the Markov blanket  are proposed for solving the problem.    
\begin{defi} \label{def4_MB}
	(Markov blanket) Let $M$ be a subset of $F$ and $f_i\in F$, ($f_i\notin M$),  then  $M$ is a Markov blanket of $f_i$ if and only if \cite{Koller_probabilistic_2009}, 
	$$f_i\perp F-{f_i}-M | M$$
\end{defi}
\begin{defi} \label{def5_Red}
	(Redundant) The feature $f_i\in F$ is said to redundant with respect to $Y$ if it has the following properties,
	\begin{itemize}
		\item[(i)] Weakly relevant to the class variable
		\item[(ii)] It has a Markov blanket $M_i$ such that $M_i\subset F-f_i$
	\end{itemize}
\end{defi}
	
\begin{figure}[h!]%
		\centering
		\includegraphics{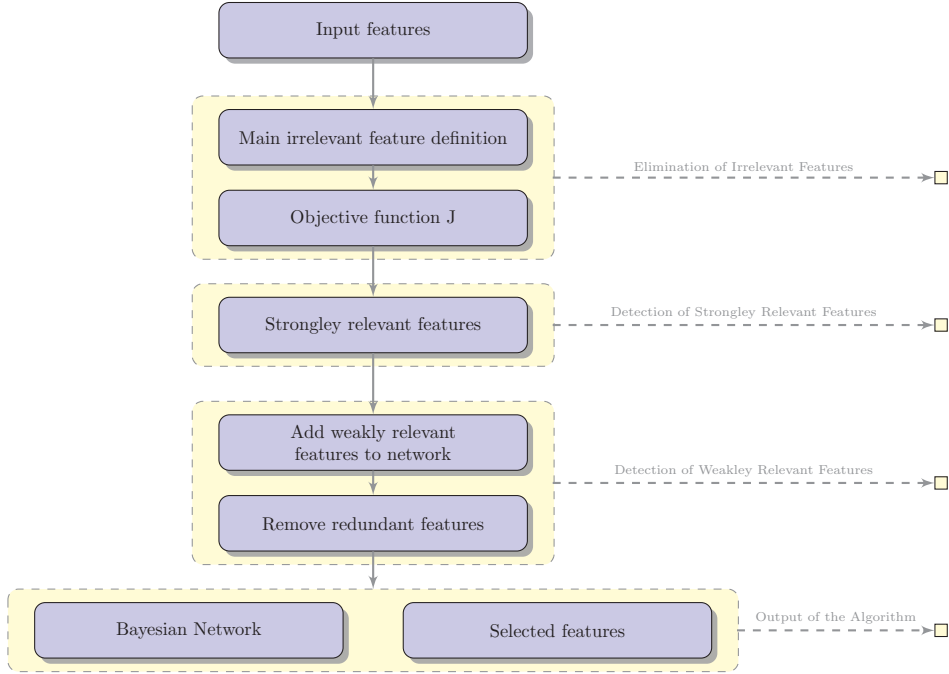}
		\caption{The overall view of the proposed algorithm}
		\label{figplan}
	\end{figure}

In conclusion, the features can be classified into four disjoint types, i) strongly relevant, ii) non-redundant weakly relevant, iii) redundant, and iv) irrelevant features. The optimal subset of features comprises strongly relevant and non-redundant weakly relevant features. Our aim is to develop a  structure learning approach to detect these two types of good features for classification tasks and provide a Bayesian network classifier generated by the selected features. The overall description of the SLFS method is shown in Figure \ref{figplan}. %

\section{The proposed Structure learning  approach for feature selection}
\label{Sec4}
Traditional machine learning techniques assume the  independence relationship among the features due to the specification complexity of  joint probability distribution among them. Probabilistic graphical models (PGM) provide a framework to assign the relationship among the features in a graph structure to  represent the distribution from it on a straightforward manner and then use the joint distribution to answer the main query based on a probabilistic approach. Although, the graph structure of  PGM should be specified by domain experts, this approach suffers from dealing with huge number of features in real applications.  This problem has attracted many researchers and a variety of techniques are proposed for identification of the dependencies among features in a graph structure,  entitled as ``Structure Learning'' \cite{murphy_machine_2012}. A variety of graph structures can be appeared based on the conditional independencies amidst features that the details of our approach for dealing with them are discussed in Sub Section \ref{SecTBN}.  We propose an integrated approach for finding the suitable structure amid features which they satisfy to the relevancy and non-redundant weakly relevant conditions.
\subsection{Structure learning based on maximum likelihood approach }
A well-known paradigm for structure learning is to apply maximum likelihood  approach as the following, 
\begin{align}
\label{eq3}
L \left( {{\theta }_{G}},G, F \right)&=\log {P}\left( F\text{ }\!\!|\!\!\text{ }{{\theta }_{G}},G \right)\cr
&=M\underset{\text{i}}{\mathop \sum }\,\hat{I}({{f}_{i}}; pa\left( {{f}_{i}} \right))-M\underset{\text{i}}{\mathop \sum }\,\hat{H}\left( {{f}_{i}} \right)
\end{align}
Where the parameters $F$, $M$, $pa(f_i)$, $\hat{H}(f_i)$,  $\hat{I}({{f}_{i}}; pa\left( {{f}_{i}} \right))$, $G$, and $\theta_G$ are the set of all features, the size of $F$,  the set of parents of $f_i$, the estimated entropy of feature $f_i$, the estimated mutual information between features $f_i$ and $pa(f_i)$, the graph structure to be learned from data, and the set of parameters to be estimated for the computation of the joint probabilities amid features. The log-likelihood is used  to simplify the calculations and for a proof of relation \eqref{eq3} one can see \cite{murphy_machine_2012}.
The identification of the redundant features requires the computation of the Markov blanket subsets of features that causes main computational complexities due to the estimation of  joint probability distributions. We have exploited Bayesian networks as the optimal graph structures amidst features because of the straightforward  identification of Markov blanket subsets through them. The main question is how to find a tree structure to discriminate between the redundant and non-redundant features. To answer this question, we propose a novel criteria  for goodness of the structure,
\begin{equation}
\label{eq4}
J = L - R
\end{equation}
where the main equation for $R$, \eqref{eq2} is approximated based on  \cite{vergara_review_2013},
\begin{align}
\label{eq5}
&|S| \approx \sum_{i}I (f_i;Y | pa(f_i),G) \cr
&I(S;Y) \approx \sum_{i}I(f_i;Y|G)
\end{align}
Hence, the relation \eqref{eq4} is changed as,
\begin{equation}
\label{eq6}
J=\sum_{i}\!\!\big(I(f_i; pa(f_i)|G) - H(f_i|G)\big)\!+\! \lambda \sum_{i}\!\!\big(I(f_i;Y | G) - I(f_i ;Y | pa(f_i),G)\big)
\end{equation}
A forward approach for simultaneous selection of optimal subset of features and building up a tree Bayesian network (TBN) according to the main criteria \eqref{eq6} is proposed as follows. The primary aim of the proposed approach for create TBN is based on the maximization of $J$ value in \eqref{eq6}. 
The TBN is built up according to the following principles,
\begin{itemize}
	\item[(i)]  The order of arrivals of input features should not be important.
	\item[(ii)] Maximize the relation \eqref{eq6}.
	\item[(iii)] Prune the features  based on Markov blanket approach for Bayesian networks
\end{itemize}
Intuitively, the principles (i) to (iii)  are given to, (i) enable our algorithm to function under incremental and online features, (ii)  consistency with the main criteria $J$, and  (iii) remove the redundant features according to Markov blanket.
\subsection{TBN construction}
\label{SecTBN}
We have assumptions on  TBN  construction regrading the main criteria \eqref{eq6}.  If there is no edge between two features in the current stage of TBN, then the mutual information between them are negligible and skipped in the computation of $J$ for the next step of the algorithm. In addition, the Markov blanket for a  feature is defined to be the set of consisting the features being in the lower depth(level) of the tree regarding it. The input feature is independent from the class given those features which have lower level in TBN based on Markov blanket definition. \\
At first, the irrelevant features identified and then removed from the remaining features. Formally $f_i$ is irrelevant if $\hat{I}\left( {{f}_{i}};Y \right)=0$. Based on our criteria \eqref{eq6}, the input feature $f_i$ is irrelevant if it decreases the value of $J$,  
which can be written as the inequalities \eqref{eq7} based on relation \eqref{eq4},
\begin{equation}
\label{eq7}
\hat{I}\left( {{f}_{i}};pa\left( {{f}_{i}} \right) \right)-\lambda\hat{I}\left( {{f}_{i}};Y\text{ }\!\!|\!\!\text{ }pa\left( {{f}_{i}} \right) \right)<0~\, \textrm{\bf and}\,\,  \lambda\hat{I}\left( {{f}_{i}};Y \right)-\hat{H}\left( {{f}_{i}} \right)<0
\end{equation}
where  $pa(f_j)$ is denoted as the parent of feature $f_i$.  The intuition behind the inequalities \eqref{eq7} comes from the definition of  irrelevant features where we expect less information between an irrelevant feature and class variable than its and other features. Moreover we set $\lambda = 1$ in this case for simplicity.\par
The TBN construction is based on weakly  and strongly relevant features that can be classified into three states.
\begin{enumerate}
	\item\label{stage2} Connect the class and input feature: When the input features increase the main relation  \eqref{eq6}, the input feature is connected to the class $Y$ and considered as a strongly relevant one. Hence connect the input feature $f_i$ to the class variable $Y$ if it holds in \eqref{eq8},
	\begin{equation}
	\label{eq8}
	\lambda \hat{I}\left( {{f}_{i}};Y \right)-H\left( {{f}_{i}} \right) >\hat{I}\left( {{f}_{i}}; pa\left( {{f}_{i}} \right) \right)-\lambda\hat{I}({{f}_{i}}; Y|pa\left( {{f}_{i}} \right))
	\end{equation}
	where the $pa(f_i)$ is selected from those nodes that directly connected to $Y$ denoted by $S$,
	\begin{equation}
	\label{eq9}
	pa({{f}_{i}})=\arg \underset{{{f}_{k}}\in S}{\mathop{\max }}\,\hat{I}\left( {{f}_{i}}; {{f}_{k}} \right)-\hat{I}({{f}_{i}}; Y|{{f}_{k}})
	\end{equation}
	The state \ref{stage2} is illustrated in Fig. \ref{fg1}.
	
	\begin{figure}[h!]%
		\centering
		\subfigure[]{	\includegraphics[width=6.5cm]{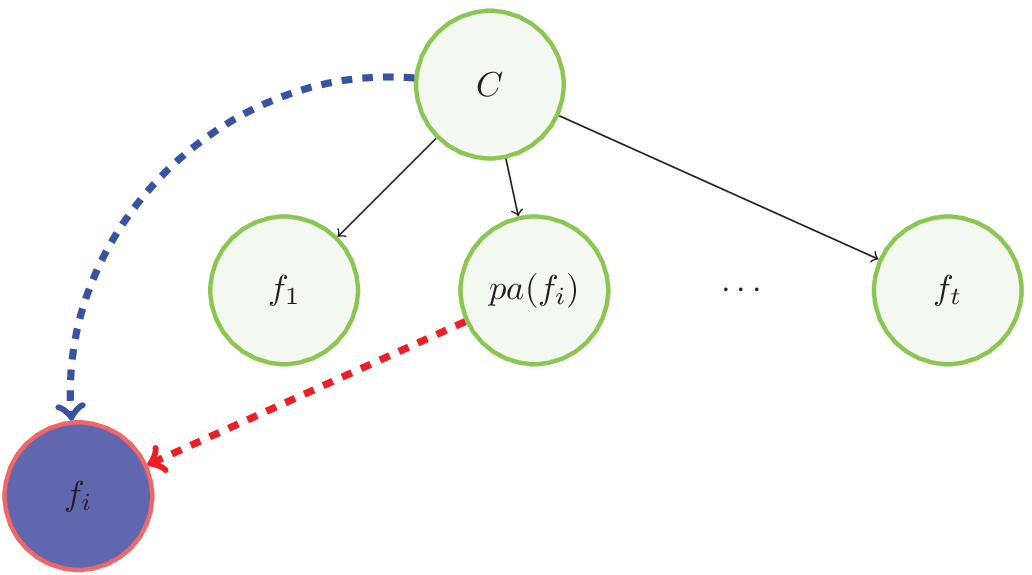}\label{fg1a}}\qquad
		\subfigure[]{ 	\includegraphics[width=6cm]{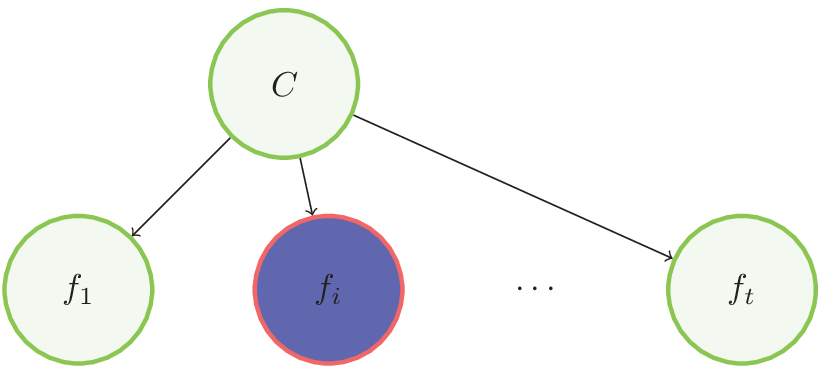}\label{fg1b}}
		\caption{Part(a): An input relevant feature $f_i$ connects to class $Y$ or a candidate parent of the TBN such as $pa(f_i)$.  Part(b):  An input feature $f_i$ satisfies in condition \eqref{eq8} which connects to $Y$.}
		\label{fg1}
	\end{figure}
	\begin{figure}[h!]%
		\centering
		\subfigure[]{	\includegraphics[width=5cm]{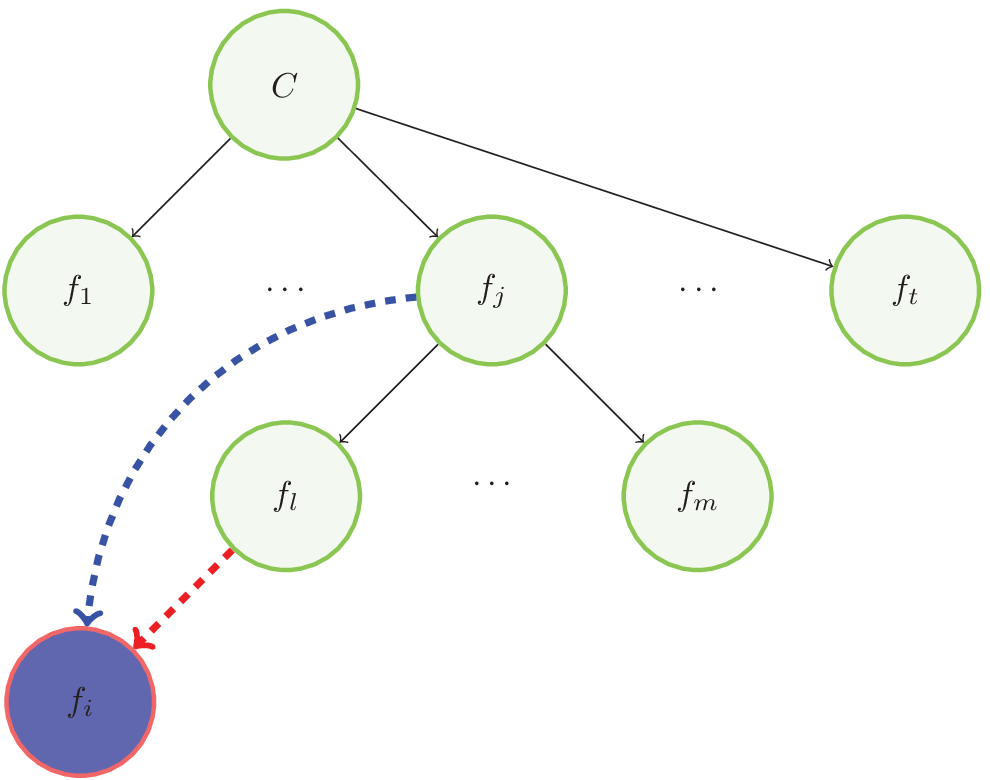}}\qquad
		\subfigure[]{ 	\includegraphics[width=5cm]{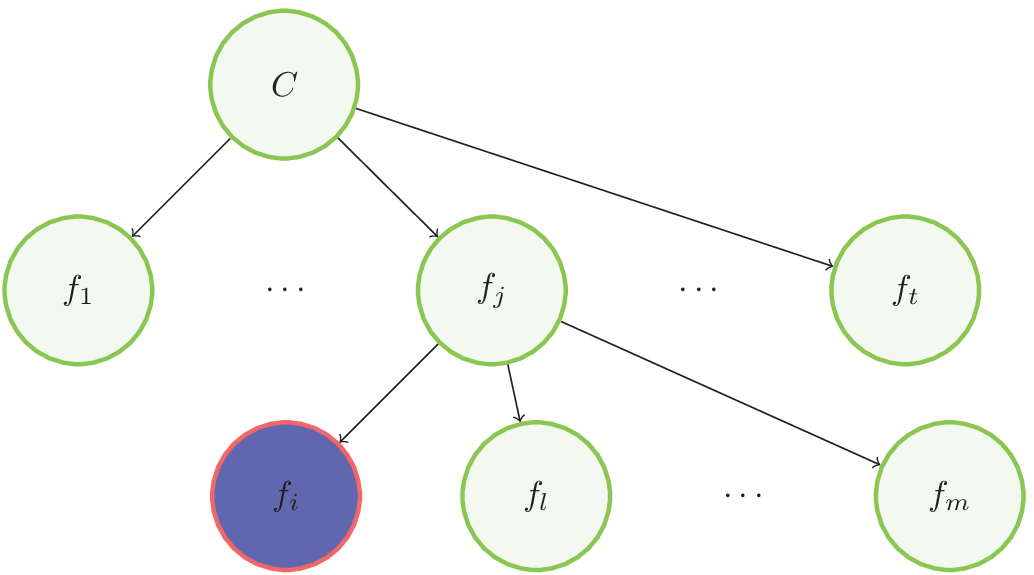}}
		\caption{Part(a): An input relevant feature $f_i$ connects to a candidate parent $f_j$ or a candidate parent between child set $f_j$, $f_l$. Part(b): An input feature $f_i$ satisfies in condition \eqref{eq10} which connects to $f_j$}
		\label{fg2}
	\end{figure}


	\item\label{stage3} Find the parent node for input feature: If the input feature $f_i$ does not satisfy in condition \eqref{eq8}, then two scenarios appear. In first scenario,  $f_i$ connects to the level (i)  of  TBN, 
	\begin{equation}
	\label{eq10}
	\hat{I}\left( {{f}_{i}};{{f}_{j}} \right)-\hat{I}\left( {{f}_{i}};Y\text{ }\!\!|\!\!\text{ }{{f}_{j}} \right)>~\hat{I}\left( {{f}_{i}};{{f}_{k}} \right)-\hat{I}({{f}_{i}};Y|{{f}_{k}})
	\end{equation}
	where the $f_j$ and $f_k$ are defined in \eqref{eq11},
	\begin{align}
	\label{eq11}
	{{f}_{j}}&=\arg \underset{{{f}_{k}}\in ch(pa(f_j))}{\mathop{\max }}\,\hat{I}\left( {{f}_{i}};{{f}_{k}} \right)-\hat{I}({{f}_{i}};Y|{{f}_{k}})\cr
	{{f}_{k}}&=\arg \underset{{{f}_{s}}\in ch(f_j)}{\mathop{\max }}\,\hat{I}\left( {{f}_{i}};{{f}_{s}} \right)-\hat{I}({{f}_{i}};Y|{{f}_{s}})
	\end{align}
	Also $f_j$ is the parent of $f_k$  and $ch(f_j)$ denotes the  child  set of $f_j$ . In second scenario, the state \ref{stage3} repeats, until the relation \eqref{eq10} is satisfied or reached to leaf in TBN.
	The Fig.\ref{fg2} describes the state \ref{stage3}.
	
\begin{figure}[h!]%
	\centering
	\subfigure[]{\includegraphics[width=5cm]{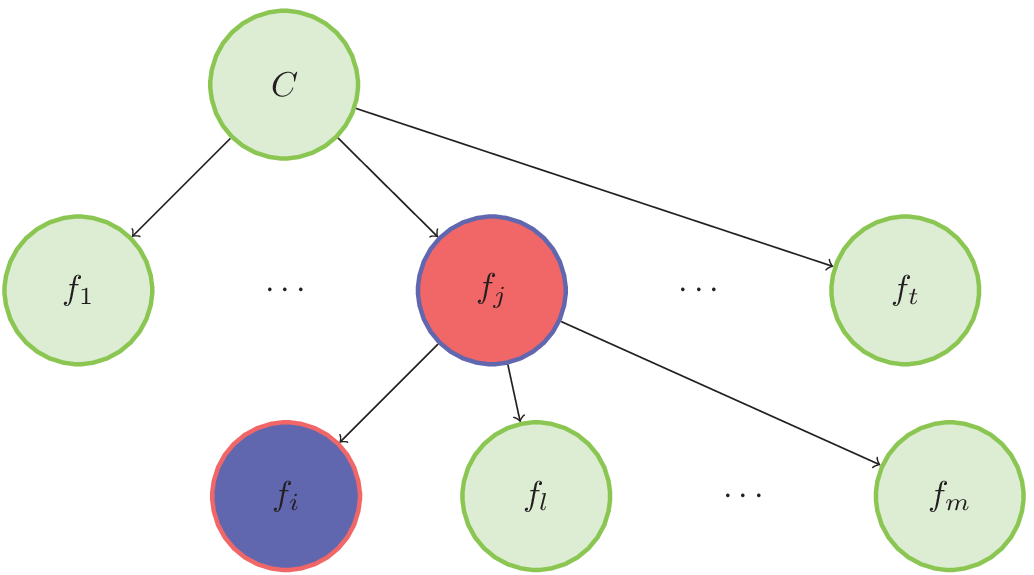}}\qquad
	\subfigure[]{ \includegraphics[width=5cm]{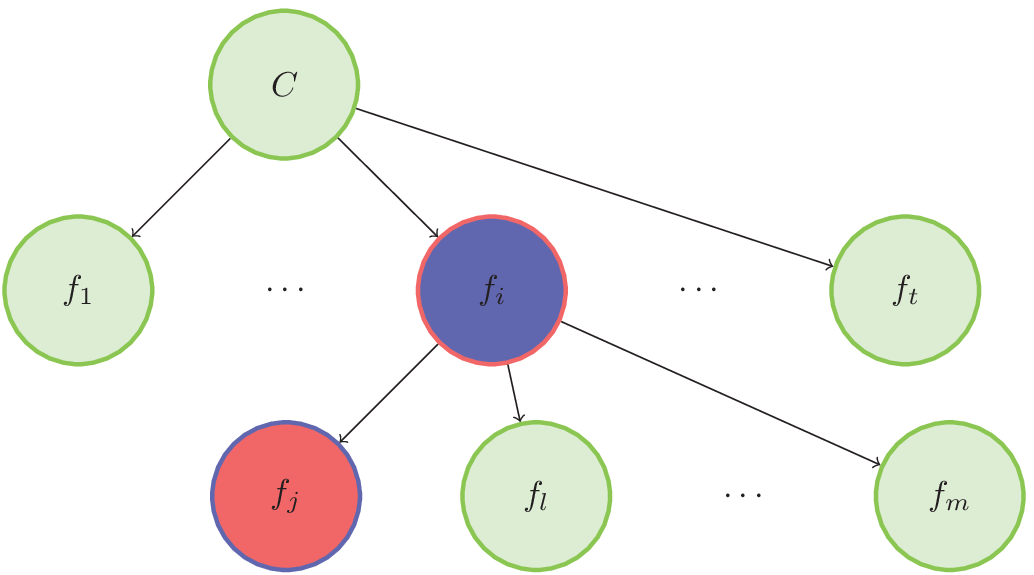}}
	\caption{Part(a): Input feature $f_i$ with state conditions \ref{stage3}. Part(b): Swap position of $f_i$ with $f_j$ based on relation \eqref{eq12}.} 
	\label{fg3}
\end{figure}
	
	\item\label{stage4} Swap input feature with the candidate parent: If  relations in \eqref{eq12} hold then swap the position of $f_i$ with $f_j$ where $f_j$ is the candidate parent selected from previous state. 
	\begin{align}
	&\hat{I}(f_i;Y) > \hat{I}(f_j;Y) \cr
	\label{eq12}
	&\hat{I}\left( {{f}_{i}};{{f}_{j}} \right)-\hat{I}\left( {{f}_{i}};Y\text{ }\!\!|\!\!\text{ }{{f}_{j}} \right)<~\hat{I}\left( {{f}_{i}};{{f}_{j}} \right)-\hat{I}({{f}_{j}};Y|{{f}_{i}})
	\end{align}
	The Fig. \ref{fg3} describes the state \eqref{stage4}. The theory behind state \ref{stage4} is given in Theorem \ref{th1}.
\end{enumerate}

\begin{figure}[h!]%
	\centering
	\subfigure[]{\label{fgt1a}	\includegraphics[width=1cm]{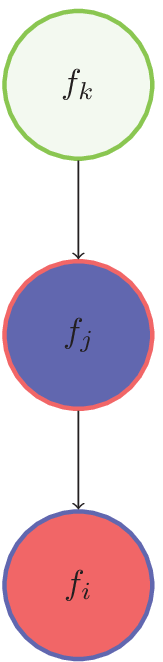}}\qquad\qquad\qquad
	\subfigure[]{\label{fgt1b} 	\includegraphics[width=1cm]{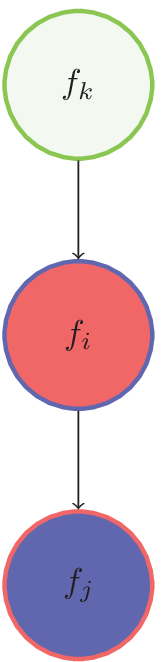}}
	\caption{The description of $f_i$, $f_j$ and $f_k$ in Theorem \ref{th1} }
	\label{fgth1}
\end{figure}

\begin{thm}
	\label{th1}
	Let $f_k , f_j$, and  $f_i$ be nodes in TBN according to Fig. \ref{fgt1a}.  If the relations \eqref{eq12} are satisfied, then the child node $f_i$ is swapped to $f_j$  to maximize the main criteria $J$ in \eqref{eq6}, (see Fig. \ref{fgt1b}). Moreover, the $f_j$ stays in this new position.
\end{thm}
\begin{pf}
The sub tree in Fig. \ref{fgt1b} is better than Fig. \ref{fgt1a} in TBN structure based on $J$ which is easily seen that from \eqref{eq12}.\\
To prove the second part of  Theorem \ref{th1}, first note that the child set of $f_j$ connects to $f_i$ based on swapping $f_i$ to $f_j$.	The following relations hold for arbitrary $f_i$, $f_j$ and $Y$ (for the proof see \cite{vergara_review_2013}), 
	\begin{equation}
	\label{eq13}
	\hat{I}\left( {{f}_{j}};Y\text{ }\!\!|\!\!\text{ }{{f}_{i}} \right)	 =
									\hat{I}\left( {{f}_{i}};Y\text{ }\!\!|\!\!\text{ }{{f}_{j}} \right)-\big(I\left( {{f}_{i}};Y \right)-\hat{I}\left( {{f}_{j}};Y \right)\big)
	\end{equation}
	Based on assumption \eqref{eq12} and relation \eqref{eq13} we have,
	\begin{equation*}
	 \hat{I}\left( {{f}_{j}};Y\text{ }\!\!|\!\!\text{ }{{f}_{i}} \right)<I({{f}_{i}};Y|{{f}_{j}})
	\end{equation*}	
	Hence it follows, 
	\begin{equation}
	\label{eq14}
	\text{ }\!\!~\!\!\text{ }\hat{I}\left( {{f}_{i}};{{f}_{j}} \right)-\hat{I}\left( {{f}_{i}};Y\text{ }\!\!|\!\!\text{ }{{f}_{j}} \right)<~\hat{I}\left( {{f}_{i}};{{f}_{j}} \right)-\hat{I}({{f}_{j}};Y|{{f}_{i}})
	\end{equation}	
We can write the relation \eqref{eq15} according to an immediate consequence of inequality \eqref{eq14},
	\begin{equation}
	\label{eq15}
	f_i=\arg \underset{\{l: f_l\in ch\{pa(f_i)\}\}}{\mathop{\max }}\,I\left( f_l; f_i \right)-I(f_j;Y|f_l)
	\end{equation}
The relation \eqref{eq15} proves the preserving the $f_j$ in the new position after swapping. 
\end{pf}
\subsection{The description of the SLFS algorithm}
The main scheme of the proposed SLFS algorithm is depicted in Algorithm \ref{alg:example}. 
\begin{algorithm}[H]
	\caption{Structure Learning for Feature Selection (SLFS)}
	\label{alg:example}
	\begin{algorithmic}[1]
		\STATE {\bfseries Input:}  $F$, $MAXDEPTH$
		\STATE {\bfseries Output:} $TBN$ comprises of the selected subset of features
		\REPEAT
		
		\IF{IRR($f_i$)}
		\STATE remove($f_i$)
		\STATE \bf{break}
		\ELSE
		\STATE $f_j$ = findParent$(f_i,childSet(Y))$
		\IF{connnect$(f_i,f_j,Y) = Y$}
		\STATE $TBN.E(Y,f_i)=1$
		\ELSE
		\STATE $depth = 1$ 
		\REPEAT
		\IF{$depth > MAXDEPTH$}
		\STATE remove($f_i$)
		\STATE {\bf{break}}
		\ENDIF
		\STATE $f_k =$ findParent$(f_i,childSet(f_j))$
		\STATE add = {\bf{false}}
		\IF{connect$(f_i,f_j,f_k) = f_j$}
		\STATE add = {\bf{true}}
		\STATE $TBN.E(f_j,f_i)=1$
		
		\ENDIF
		
		\IF{swapCheck$(f_i,f_j)$}
		\STATE TBN.swap$(f_i,f_j)$
		\ENDIF
		\STATE {$depth = depth+1$}
		\UNTIL{add = {\bf{false}}}
		\ENDIF
		\ENDIF
		\UNTIL{$\forall \ f_i \ \in \ F$}
	\end{algorithmic}
\end{algorithm}
The Algorithm \ref{alg:example} initially depends on the set of current available features $F$,  maximum level of the tree $MAXDEPTH$, and the regularization parameter $\lambda$. 
In this algorithm first we check the irrelevant features with \emph{IRR } function depending on the $\lambda$ based on relation \eqref{eq7}. Lines 6-29 explain the different situation for adding features according to State \ref{stage2} to State \ref{stage4}. The \emph{findParent} function finds a candidate parent for input feature $f_i$ according to relation \eqref{eq9}. The \emph{connect} function select a suitable parent for $f_i$ which maximize $J$ based on state conditions \ref{stage2} and \ref{stage3}. Adding edge to TBN graph is performed by \emph{TBN.E}. Function \emph{swapCheck}  checks the correctness of condition of swapping the position between two features in TBN based on State \ref{stage4}. The swap position between features is done by \emph{swap} function. Lines 19-22 and lines 23-25 perform based upon  State \ref{stage3} and State \ref{stage4}.
If a feature goes down the maximum depth of the TBN, \emph{MAXDEPTH} according to the algorithm procedure, it is removed from the selected subset of features. Moreover,  the features in the final TBN are considered as the Markov blanket of  the removed features for the output variable $Y$.

\begin{table}[t!]
	\centering
	\caption{The time complexity of the main used algorithms, where $p$, $N$,  $|S|$ and $K$ are the number of features,  number of samples,   size of selected features and maximum allowable size of selected candidate subset of features.}
	\label{tb1}
	\begin{tabular}{lc}
		\toprule[1.5pt]
		Algorithm & Time Complexity\\
		\midrule 
		SLFS & $O(Npd^2)$\\
		MRMR \cite{peng_MRmr_2005}& $O(Np|S|)$\\
		fast-OSFS \cite{wu_osfs_2013} &$O(Np |S| K^{|S|})$\\
		alpha-investing \cite{zhou2005streaming_alphainvesting}& $O(Np |S|^2)$\\
		chi-square \cite{liu1995chi2} & $O(Np)$\\
		\bottomrule[1.5pt]
		
	\end{tabular} 
\end{table}

\subsection{Time complexity analysis}
We compare the SLFS algorithm to the earlier algorithms based on the time complexity. A summary of  time complexity analysis is presented in Table \ref{tb1}. The main computational part of SLFS  is the step of feature adding in TBN construction. The feature adding step consists up two elements. Comparison operation of the given feature with the other existing features in each depth level is done in first element. Second one is the calculation of mutual information between two features. If we assume that the maximum child for each existing node and the total levels of the TBN  equal to \emph{NCH} and \emph{MAXDEPTH}, then the total number of comparison is at most $\textrm{\emph{NCH}}\times\textrm{\emph{MAXDEPTH}}$. So, the time complexity of the first element is at most $O(p)$ for each feature with the assumption  $\textrm{\emph{MAXDEPTH}}\ll p$ and $\textrm{\emph{NCH}}\longrightarrow p$.
The time complexity of the second element is at most $O(Nd^2)$ where $d$ is the number of distinct values of a discrete features.   Hence, the overall computational complexity of the proposed method is $O(Np^2d^2)$ where $N$ is the number of samples. The practical assumptions on limits  $\textrm{\emph{MAXDEPTH}}\leq 5$ and $\textrm{\emph{NCH}}\leq 15$ reduce the time complexity of SLFS to $O(Npd^2)$.
In fact,  the SLFS benefits from the optimum time complexity versus the other works such as \emph{fast osfs}  \cite{wu_osfs_2013}, \emph{mrmr} \cite{peng_MRmr_2005}, and \emph{alpha-investing} \cite{zhou2005streaming_alphainvesting} based on relation $d<|S|<K$. While the computational complexity of the chi-square based method \cite{liu1995chi2} is less than the SLFS algorithm, it suffers from the greedy ranking based approach.  
The proposed algorithm takes advantage  the independency of learning algorithms in the feature selection process. In addition, the computational complexity of the SLFS method is much less than the wrapper-based evaluation techniques. 
\section{Experiments}
\label{Sec5}
In this section we provide the experimental evaluation of the proposed approach versus the other well-known methods through a variety of different frequently used datasets. 
\subsection{Datasets description}
\begin{table}[t!]
	\centering
	\caption{The description of the benchmark datasets used in our study, where $N$, and $p$ denote the number of samples, and number of features}
	\label{tbdataset}
	\begin{tabular}{lccc}
		\toprule[1pt]
		Data sets & $N$ & $p$ & $Classes$\\
		\midrule  
		ARCENE & 200 & 10000 & 2\\
		BreastCancer & 683 & 9 & 2\\
		Dexter & 600 & 20000 & 2\\
		Dorothea & 1150 & 100000 & 2\\
		Isolet & 7797 & 617 & 26\\
		Madelon & 2600 & 500 & 2\\
		Voting & 435 & 16 & 2\\
		Yeast &       1484  &    8  &  10\\
		Letter&        20000 &    16   &  28 \\
	\end{tabular} 
\end{table}

To evaluate the proposed SLFS method, we have used a bunch of benchmark datasets which have been applied in many works such as \cite{moradi_integration_2015,wu_osfs_2013, peng_MRmr_2005, Liu04-jmlr, Moradi_relevanceredundancy_2015}. The different datasets include small to large number of features and observations in two-category or multi-category classification problems. A summary of the datasets are given in Table \ref{tbdataset} available online from UCI repository \cite{uciLichman2013}. 
ARCENE  is a two-class classification dataset with continuous input variables to distinguish cancer versus normal patterns.
BreastCancer data samples consist of visually assessed nuclear features of fine needle aspirates (FNAs) taken from patients' breasts. The aim is to  predict the presence or absence of a malignant tumor from the FNA results based on real-valued input variables. 
Dexter is  a two-class  dataset with sparse continuous input variables to filter text corpus in ``corporate acquisitions''.  Dorothea is a drug discovery dataset where chemical compounds represented by structural molecular features must be classified as active (binding to thrombin) or inactive. 
Isolet dataset contains 150 subjects who spoke the name of each letter of the alphabet twice. The aim of this study was to predict  which letter-name was spoken. 
Madelon is a two-category classification  dataset presented in the feature selection challenge of NIPS 2003.  This dataset suffers from the high dimensionality of number of  features and samples  simultaneously. Because of the mixing the data  by adding noise, flipping labels, shifting and rescaling, it resulted in a hard dataset for feature selection task. 
The Voting data comprises voting information for 435 samples where each one is about a person voting  on 16 issues. The aim is to classify a person as  republican or democrat based on these categorical features. 
Yeast dataset contains information about  a set of Yeast cells.  The target is to specify the localization site of each cell among 10 possible alternatives.
Letter is a dataset of $20000$ black and white rectangular pixel displays of one of the 26 capital letters of English alphabet. The aim is to recognize the right letter for unseen observations of images.
Most of these datasets are presented in feature selection challenge held in NIPS  \cite{guyon2004result}.

\subsection{The experimental setting}
\begin{table}[t!]
	\centering
	\caption{The sensitivity analysis of SLFS algorithm to its parameters for Isolet dataset, where NCH, lambda, MAXDEPTH and NSF are largest number of child for each node, Value of $\lambda$ in primary criteria \eqref{eq6}, the depth cut  and the number of selected features.}
	\label{tbIsolet}
	\begin{tabular}{ccccc}
		\toprule[1pt]
		NCH & lambda & MAXDEPTH & NSF & accuracy\\
		\midrule
		15 & 1 & 2 & 80 & 77.51\\
		8 & 1.5 & 2 & 52 & 62.43\\
		8 & 1.5 & 3 & 194 & 33.23\\
		15 & 1.5 & 2 & 84 & 62.94\\
		15 & 0.75 & 1 & 16 & 75.64\\
		15 & 1 & 1 & 16 & 82.07\\
	\end{tabular} 
\end{table}

We compare our algorithm with the well-known methods, MRMR \cite{peng_MRmr_2005}, OSFS \cite{wu_osfs_2013}, Alpha-investing \cite{zhou2005streaming_alphainvesting} and Chi-Square \cite{liu1995chi2} based on prediction accuracy. To compare the performance of proposed method with other feature selection algorithms  $10$-fold cross-validation is used in our experiments. 
The discretization step of the proposed approach is performed based on  \cite{fayyad_multi-interval_1993}. \par
We  perform a case study on Isolet dataset with three different \emph{MAXDEPTH} values to sensitivity analysis of the proposed approach on the \emph{MAXDEPTH} parameter.  Based on  results in Table \ref{tbIsolet}, the SLFS method is less sensitive to this parameter as a cut depth in TBN construction. \par
Because of the search strategy of the  SLFS method is independent of the classification algorithms,  and so we expect that the proposed approach meet the good prediction accuracy based on the different classifiers. For this aim, three well-known classification algorithms, support vector machine (SVM) \cite{cortes_support-vector_1995},  k nearest neighbor (KNN) \cite{bishop_pattern_2007}, and  Naive Bayes (NB) \cite{Hastie_elements_2011} have been applied to test the algorithms. 
\begin{figure}[h!]%
	\centering
	\includegraphics[scale = 1, width = 12cm]{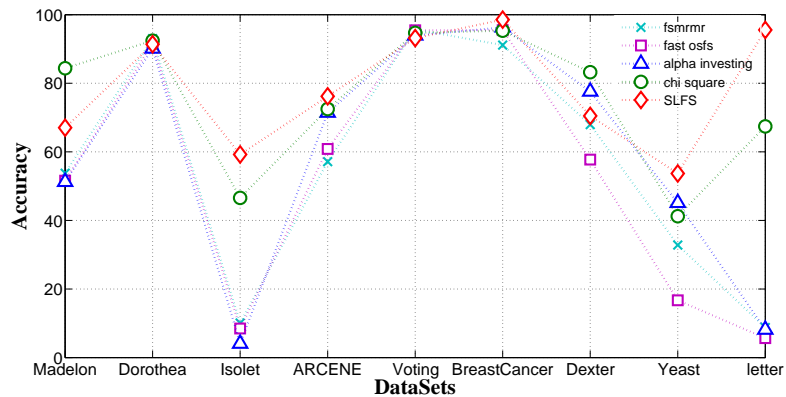}
	\caption{The prediction accuracy with KNN (K = 3) classifier on  selected subset of features by the specified algorithms}
	\label{fgKNN3}
\end{figure}
\begin{figure}[h!]%
	\centering
	\includegraphics[scale = 1, width = 12cm]{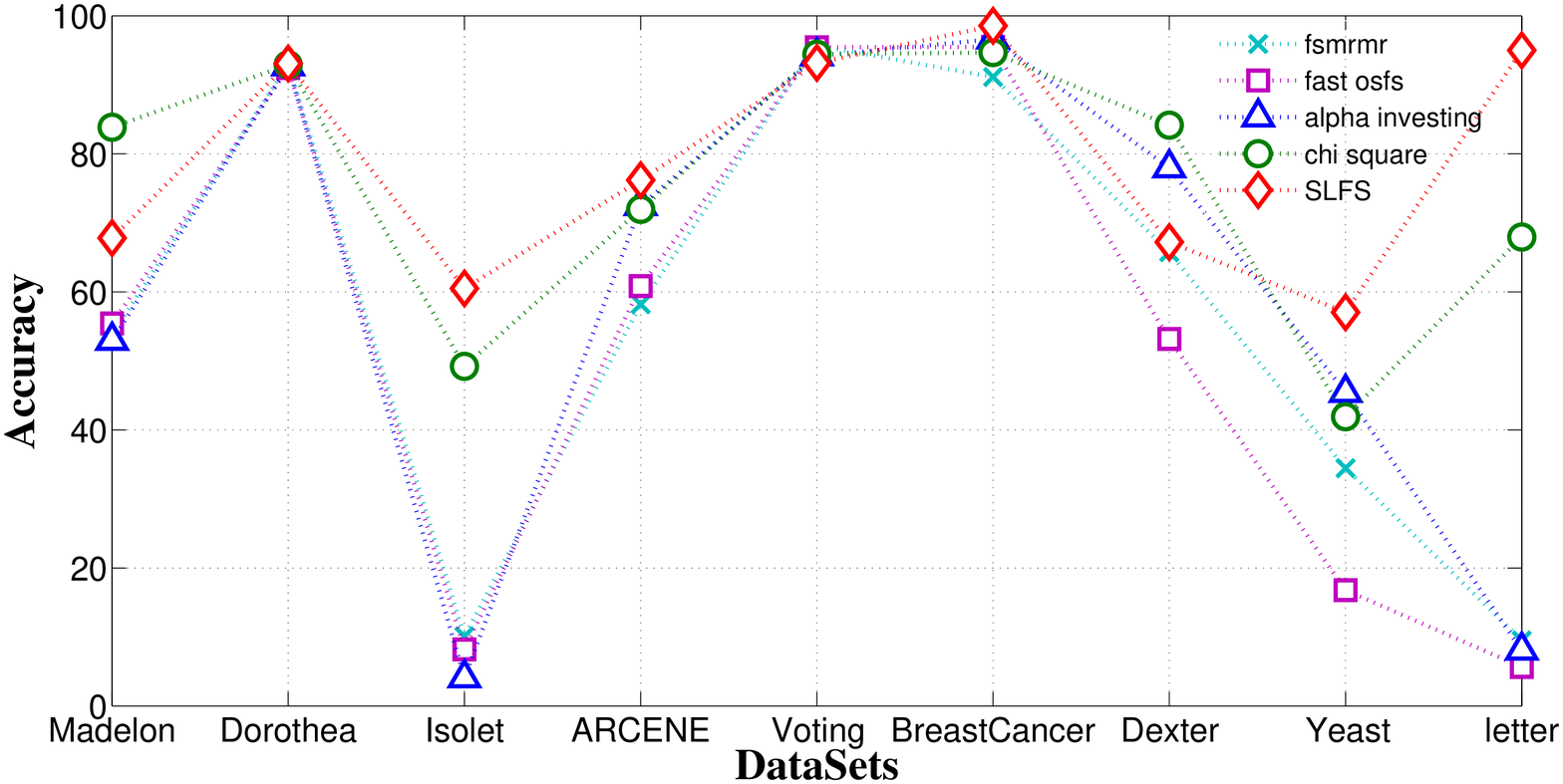}
	\caption{The prediction  accuracy with KNN (K = 5) classifier on subset of features by the specified algorithms}
	\label{fgKNN5}
\end{figure}
\begin{figure}[h!]%
	\centering
	\includegraphics[scale=1,width=12cm]{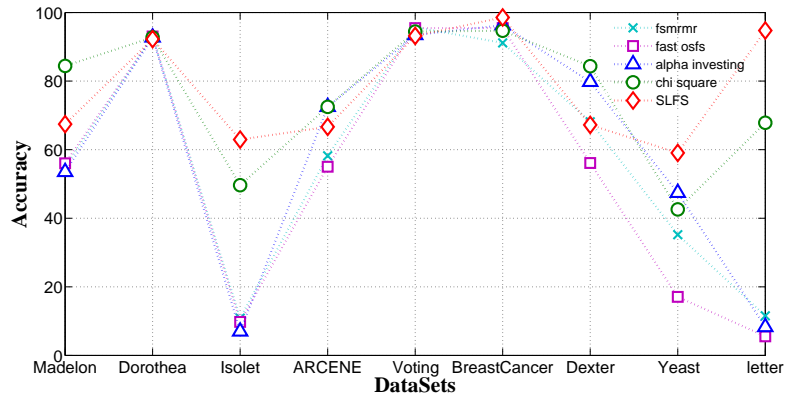}
	\caption{The prediction accuracy with KNN (K = 7) classifier on subset of features by the specified algorithms}
	\label{fgKNN7}
\end{figure}
\subsection{Experimental results}
\begin{figure}[h!]%
	\centering
	\includegraphics[scale=1,width=12cm]{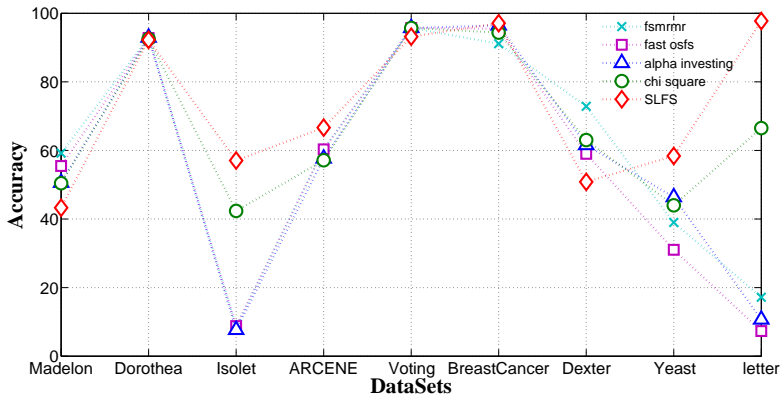}
	\caption{The prediction accuracy with SVM classifier on subset of features by the specified algorithms}
	\label{fgSVM}
\end{figure}

\begin{figure}[h!]%
	\centering
	\includegraphics[scale=1,width=12cm]{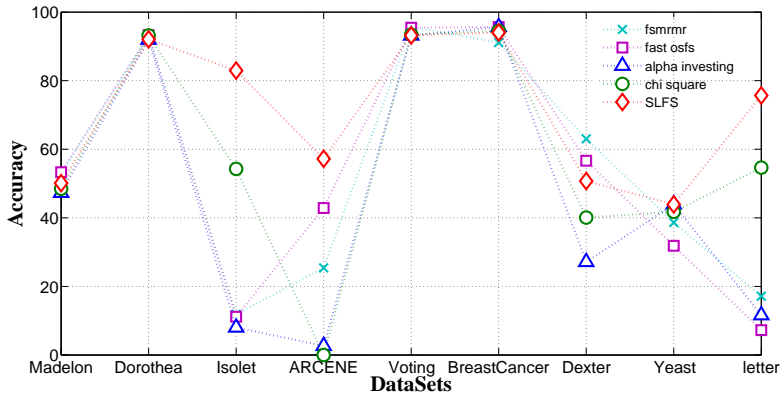}
	\caption{The prediction accuracy  comparison between the Naive Bayes classifier on selected subset of features of the other specified  algorithms and  BNSLFS classifier on selected features obtained by our approach}
	\label{fgSLFS}
\end{figure}

We apply KNN and SVM classifiers on the selected subset of features and report the average prediction accuracy in the performed experiments.  In line with the two used classifier on the selected subset of features in Fig. \ref{fgKNN3} to Fig \ref{fgSVM}, we compare our induced Bayesian network classifier, hereafter called as \emph{``BNSLFS''}, with  NB  applied on the  selected subset of features  from the other feature selection approaches presented in Fig. \ref{fgSLFS}. 
The prediction accuracy, the number of correctly classified samples over the total number of samples, is used as the performance measure in the experiments. 
We have used 10-fold cross validation techniques to get the average reported prediction accuracy.
In the plots, the x-axis denotes the dataset, and the y-axis denotes the average prediction  accuracy  based on 10-fold cross validation  on the selected subset of features by the specified feature selection algorithm. 

Fig. \ref{fgKNN3} to Fig. \ref{fgKNN7} show the results of the KNN based classifier with $K = 3, 5, 7$ run on the applied feature selection techniques. Fig. \ref{fgKNN3} shows the prediction accuracy of our method is better than the others based on  3-NN classifier on the selected subset of features in the \emph{BreastCancer}, \emph{Isolet}, \emph{Voting},  \emph{Yeast} and  \emph{Letter} datasets. In addition, the SLFS is performed equally or slightly weaker than the other methods based on \emph{ARCENE}, \emph{Madelon}, \emph{Dorothea} and  \emph{Dexter} in Fig. \ref{fgKNN3}.\par
 The results of the 5-NN classifier in Fig. \ref{fgKNN5} based on selected features of SLFS  show superior or just as the other feature selection algorithms. In addition,  the Fig. \ref{fgKNN7} shows that the SLFS  based 7-NN classifier is superior to the other methods on the \emph{Isolete}, \emph{Voting}, \emph{BreastCancer}, \emph{Dexter}, \emph{Yeast} and  \emph{Letter} dataset. In addition, the SLFS based 7-NN results are slightly weaker than the fast-osfs and fsmrmr methods based classifier on the  \emph{Madelon} dataset.\par
Fig. \ref{fgSVM} shows the results of SVM classifier based on the feature selection techniques.  We have used SVM classifier based on a linear polynomial kernel form LIBSVM software \cite{svmlib}. We can observe that  the SVM based on SLFS method, performs better than the other SVM based feature selection techniques in \emph{BreastCancer}, \emph{Isolet}, \emph{Voting}, and \emph{Dexter}, \emph{Yeast} and  \emph{Letter}. Furthermore, the SLFS is performed equally or slightly weaker than the other methods based on \emph{ARCENE}, \emph{Madelon}, \emph{Dorothea} in Fig. \ref{fgSVM}.\par

In line with the aim of feature selection with the SLFS algorithm,  the SLFS  provides us an induced Bayesian network classifier (BNSLFS) on the training dataset. To compare with the other methods, the Naive Bayes is used for the other feature selection techniques. The results are presented in Fig. \ref{fgSLFS}. These results show us the superiority of  the \emph{BNSLFS} than the other techniques for the \emph{ARCENE}, \emph{Isolet}, \emph{Yeast} and  \emph{Letter} datasets.\par The earlier results in Fig. \ref{fgKNN3} through Fig. \ref{fgSVM} based on KNN and SVM classifiers showed  us that the best performance of the classifiers based on our feature selection approach  have been occurred on the multi-category datasets.  Moreover, the BNSLFS is significantly better than the other techniques on the \emph{Isolet} dataset comprising very noisy data that confirms the strength of graphical structure learning among the features through the process of SLFS method.
Not only the BNSLFS classifier benefits from the no need of the added cost of classifier training on the selected subset of features, but  its performance  also shows reasonable accuracy as compared with the other well-known methods.
\begin{figure}[htp!]%
	\centering
	\subfigure[]{\label{fgBN1a}	\includegraphics[width=7cm]{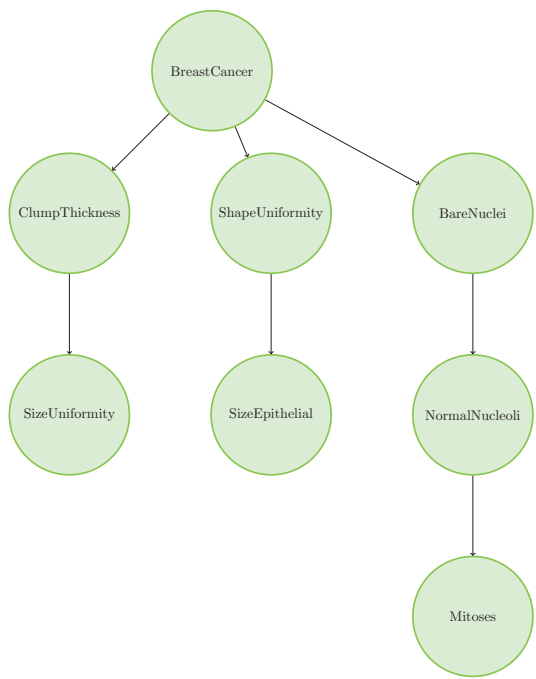}}\qquad\qquad\qquad
	\subfigure[]{\label{fgBN1b} \includegraphics[width=8.5cm]{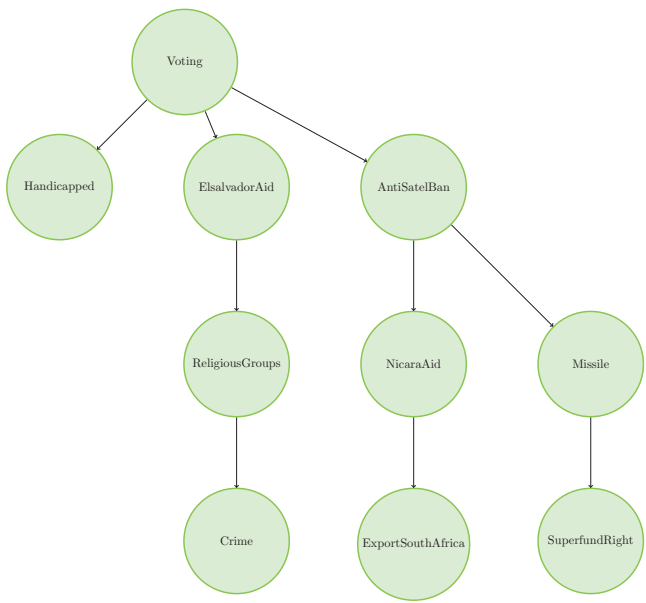}}
	\caption{The Bayesian networks of two datasets, \emph{BreastCancer} in \textbf{(a)} and \emph{Voting} in \textbf{(b)} with MAXDEPTH = 4.}
	\label{fgBN}
\end{figure}

The Bayesian network structures can be used for knowledge discovery and density estimation tasks where the aim of the former is to discover the complex structure among features and the primary aim of the latter is to discover the features relationships to compute the joint distributions among them for learning tasks such as classification and model based clustering \cite{murphy_machine_2012}. While in this study we focused on the structure learning for feature selection and classification aims, the proposed method can be applied for knowledge discovery task such as understanding the complex phenomenon including the genes and DNA arrays. Here, we present two graph structures for the ``Voting'' and ``BreastCancer'' datasets in Figure \ref{fgBN}. The obtained Bayesian network for these datasets demonstrates the relationship among the selected features, and  the strength of effects of them on the class variable. In line with the reduction of high dimensional dataset with huge number of features in the SLFS procedure, it can enable us to apply the Bayesian network  for knowledge discovery and structure representation for illustration the complex phenomena.
\section{Conclusion}
\label{Sec6}
In this paper we have developed a novel feature selection technique based on integrating the structure learning and the Markov blanket optimal theoretical concept.
While the typical feature selection methods have suffered from the greedy or pairwise methods for distinguishing between the redundant and non-redundant features, the SLFS approach allowed us to identify redundant features with the aid of TBN's.
In line with the theoretical works for deriving the SLFS algorithm, the optimality of the selected features based on this approach was presented through a variety of benchmark datasets. The experimental results based on SVM and KNN classifiers trained on the selected subset of features,  showed us the superiority of prediction accuracy of classification through the SLFS feature selection algorithm as compared with the other  feature selection methods.
In line with a better local optimum set of features according to the proposed approach rather than the other  techniques, our feature selection approach was provided a Bayesian network classifier  without the additional cost of classifier training on the selected features dissimilar to the typical supervised feature selection methods. Furthermore, the \emph{BNSLFS} performance  on the benchmark dataset  showed  the better or equal accuracy of  it versus the Naive Bayes classifier on the selected  features according to other feature selection methods.\\ 
There exist suggestions for future works  to extend this research such as follows, 
\begin{itemize}
	\item Statistical significance of the selected features of the SLFS algorithm based on statistical hypothesis tests
	\item  The comparison of the proposed approach with the other structure learning techniques such as Tree augmented Naive Bayes (TAN)
	\item Generalization of the other feature selection techniques through the framework of  probabilistic graphical models 
\end{itemize}
\section*{References}
\bibliographystyle{elsarticle-num} 
\bibliography{SLFS.bib}

\begin{thebibliography}{10}
\expandafter\ifx\csname url\endcsname\relax
  \def\url#1{\texttt{#1}}\fi
\expandafter\ifx\csname urlprefix\endcsname\relax\def\urlprefix{URL }\fi
\expandafter\ifx\csname href\endcsname\relax
  \def\href#1#2{#2} \def\path#1{#1}\fi

\bibitem{wang_onlineFS_2014}
J.~Wang, P.~Zhao, S.~Hoi, R.~Jin, Online {Feature} {Selection} and {Its}
  {Applications}, IEEE Transactions on Knowledge and Data Engineering 26~(3)
  (2014) 698--710.
\newblock \href {http://dx.doi.org/10.1109/TKDE.2013.32}
  {\path{doi:10.1109/TKDE.2013.32}}.

\bibitem{tan_towardsUltraHigh_2014}
M.~Tan, I.~W. Tsang, L.~Wang, Towards {Ultrahigh} {Dimensional} {Feature}
  {Selection} for {Big} {Data}, J. Mach. Learn. Res. 15~(1) (2014) 1371--1429.

\bibitem{bishop_pattern_2007}
C.~M. Bishop, Pattern {Recognition} and {Machine} {Learning}, 1st Edition,
  Springer, 2007.

\bibitem{ida13}
H.~Zare, A.~Mohammadpour, P.~Moradi,
  \href{http://dx.doi.org/10.3233/IDA-130575}{A random projection approach for
  estimation of the betweenness centrality measure}, Intell. Data Anal. 17~(2)
  (2013) 217--231.
\newblock \href {http://dx.doi.org/10.3233/IDA-130575}
  {\path{doi:10.3233/IDA-130575}}.
\newline\urlprefix\url{http://dx.doi.org/10.3233/IDA-130575}

\bibitem{Moradi_relevanceredundancy_2015}
S.~Tabakhi, P.~Moradi, Relevance--redundancy feature selection based on ant
  colony optimization, Pattern Recognition 48~(9) (2015) 2798--2811.
\newblock \href {http://dx.doi.org/10.1016/j.patcog.2015.03.020}
  {\path{doi:10.1016/j.patcog.2015.03.020}}.

\bibitem{peng_MRmr_2005}
H.~Peng, F.~Long, C.~Ding, Feature selection based on mutual information
  criteria of max-dependency, max-relevance, and min-redundancy, IEEE
  Transactions on Pattern Analysis and Machine Intelligence 27~(8) (2005)
  1226--1238.
\newblock \href {http://dx.doi.org/10.1109/TPAMI.2005.159}
  {\path{doi:10.1109/TPAMI.2005.159}}.

\bibitem{wu_osfs_2013}
X.~Wu, K.~Yu, W.~Ding, H.~Wang, X.~Zhu, Online {Feature} {Selection} with
  {Streaming} {Features}, IEEE Transactions on Pattern Analysis and Machine
  Intelligence 35~(5) (2013) 1178--1192.
\newblock \href {http://dx.doi.org/10.1109/TPAMI.2012.197}
  {\path{doi:10.1109/TPAMI.2012.197}}.

\bibitem{zokaei_ashtiani_bandit-based_2014}
M.-H. Zokaei~Ashtiani, M.~Nili~Ahmadabadi, B.~Nadjar~Araabi,
  \href{http://www.sciencedirect.com/science/article/pii/S0925231214002173}{Bandit-based
  local feature subset selection}, Neurocomputing 138 (2014) 371--382.
\newblock \href {http://dx.doi.org/10.1016/j.neucom.2014.02.001}
  {\path{doi:10.1016/j.neucom.2014.02.001}}.
\newline\urlprefix\url{http://www.sciencedirect.com/science/article/pii/S0925231214002173}

\bibitem{wang_feature_2015}
D.~Wang, F.~Nie, H.~Huang, Feature {Selection} via {Global} {Redundancy}
  {Minimization}, IEEE Transactions on Knowledge and Data Engineering 27~(10)
  (2015) 2743--2755.
\newblock \href {http://dx.doi.org/10.1109/TKDE.2015.2426703}
  {\path{doi:10.1109/TKDE.2015.2426703}}.

\bibitem{murphy_machine_2012}
K.~Murphy, Machine {Learning}: {A} {Probabilistic} {Perspective}, 1st Edition,
  The MIT Press, Cambridge, Mass., 2012.

\bibitem{feng_unsupervised_2016}
J.~Feng, L.~Jiao, F.~Liu, T.~Sun, X.~Zhang,
  \href{http://www.sciencedirect.com/science/article/pii/S0031320315003064}{Unsupervised
  feature selection based on maximum information and minimum redundancy for
  hyperspectral images}, Pattern Recognition 51 (2016) 295--309.
\newblock \href {http://dx.doi.org/10.1016/j.patcog.2015.08.018}
  {\path{doi:10.1016/j.patcog.2015.08.018}}.
\newline\urlprefix\url{http://www.sciencedirect.com/science/article/pii/S0031320315003064}

\bibitem{moradi_graphUnsFS_2015}
P.~Moradi, M.~Rostami, A graph theoretic approach for unsupervised feature
  selection, Engineering Applications of Artificial Intelligence 44 (2015)
  33--45.
\newblock \href {http://dx.doi.org/10.1016/j.engappai.2015.05.005}
  {\path{doi:10.1016/j.engappai.2015.05.005}}.

\bibitem{tabakhi_unsupervised_2014}
S.~Tabakhi, P.~Moradi, F.~Akhlaghian, An unsupervised feature selection
  algorithm based on ant colony optimization, Engineering Applications of
  Artificial Intelligence 32 (2014) 112--123.
\newblock \href {http://dx.doi.org/10.1016/j.engappai.2014.03.007}
  {\path{doi:10.1016/j.engappai.2014.03.007}}.

\bibitem{perkins_online_2003}
S.~Perkins, J.~Theiler,
  \href{http://www.aaai.org/Library/ICML/2003/icml03-078.php}{Online feature
  selection using grafting}, in: Machine Learning, Proceedings of the Twentieth
  International Conference {(ICML} 2003), August 21-24, 2003, Washington, DC,
  {USA}, 2003, pp. 592--599.
\newline\urlprefix\url{http://www.aaai.org/Library/ICML/2003/icml03-078.php}

\bibitem{liu_toward_2005}
H.~Liu, L.~Yu, Toward {Integrating} {Feature} {Selection} {Algorithms} for
  {Classification} and {Clustering}, IEEE Trans. Knowl. Data Eng. 17~(4) (2005)
  491--502.
\newblock \href {http://dx.doi.org/10.1109/TKDE.2005.66}
  {\path{doi:10.1109/TKDE.2005.66}}.

\bibitem{guyon_introduction_2003}
I.~Guyon, A.~Elisseeff, An introduction to variable and feature selection, The
  Journal of Machine Learning Research 3 (2003) 1157--1182.

\bibitem{DashLiu97-ida}
M.~Dash, H.~Liu, Feature selection for classification, Intell. Data Anal.
  1~(1-4) (1997) 131--156.
\newblock \href {http://dx.doi.org/10.1016/S1088-467X(97)00008-5}
  {\path{doi:10.1016/S1088-467X(97)00008-5}}.

\bibitem{liu_feature_2015}
Y.~Liu, F.~Tang, Z.~Zeng, Feature {Selection} {Based} on {Dependency} {Margin},
  IEEE Transactions on Cybernetics 45~(6) (2015) 1209--1221.
\newblock \href {http://dx.doi.org/10.1109/TCYB.2014.2347372}
  {\path{doi:10.1109/TCYB.2014.2347372}}.

\bibitem{KohaviJ97-ai}
R.~Kohavi, G.~H. John, Wrappers for feature subset selection, Artif. Intell.
  97~(1-2) (1997) 273--324.
\newblock \href {http://dx.doi.org/10.1016/S0004-3702(97)00043-X}
  {\path{doi:10.1016/S0004-3702(97)00043-X}}.

\bibitem{MajiG13-tcyb}
P.~Maji, P.~Garai, Fuzzy-rough simultaneous attribute selection and feature
  extraction algorithm, {IEEE} T. Cybernetics 43~(4) (2013) 1166--1177.
\newblock \href {http://dx.doi.org/10.1109/TSMCB.2012.2225832}
  {\path{doi:10.1109/TSMCB.2012.2225832}}.

\bibitem{moradi_integration_2015}
P.~Moradi, M.~Rostami, Integration of graph clustering with ant colony
  optimization for feature selection, Knowledge-Based Systems 84 (2015)
  144--161.

\bibitem{BressanV03-pami}
M.~Bressan, J.~Vitri{\`{a}}, On the selection and classification of independent
  features, {IEEE} Trans. Pattern Anal. Mach. Intell. 25~(10) (2003)
  1312--1317.
\newblock \href {http://dx.doi.org/10.1109/TPAMI.2003.1233904}
  {\path{doi:10.1109/TPAMI.2003.1233904}}.

\bibitem{OliveiraS92-icml}
A.~L. Oliveira, A.~L. Sangiovanni{-}Vincentelli, Constructive induction using a
  non-greedy strategy for feature selection, in: Proceedings of the Ninth
  International Workshop on Machine Learning {(ML} 1992), Aberdeen, Scotland,
  UK, July 1-3, 1992, 1992, pp. 355--360.

\bibitem{Lewis92}
D.~D. Lewis, Feature selection and feature extraction for text categorization,
  in: Proceedings of the Workshop on Speech and Natural Language, HLT '91,
  Association for Computational Linguistics, Stroudsburg, PA, USA, 1992, pp.
  212--217.
\newblock \href {http://dx.doi.org/10.3115/1075527.1075574}
  {\path{doi:10.3115/1075527.1075574}}.

\bibitem{Geng07--FSRanking}
X.~Geng, T.-Y. Liu, T.~Qin, H.~Li, Feature selection for ranking, in:
  Proceedings of the 30th Annual International ACM SIGIR Conference on Research
  and Development in Information Retrieval, SIGIR '07, ACM, New York, NY, USA,
  2007, pp. 407--414.
\newblock \href {http://dx.doi.org/10.1145/1277741.1277811}
  {\path{doi:10.1145/1277741.1277811}}.

\bibitem{PinheiroCCR12-eswa}
R.~H.~W. Pinheiro, G.~D.~C. Cavalcanti, R.~F. Correa, T.~I. Ren, A
  global-ranking local feature selection method for text categorization, Expert
  Syst. Appl. 39~(17) (2012) 12851--12857.
\newblock \href {http://dx.doi.org/10.1016/j.eswa.2012.05.008}
  {\path{doi:10.1016/j.eswa.2012.05.008}}.

\bibitem{Liu1998-Book}
H.~Liu, H.~Motoda, Feature Selection for Knowledge Discovery and Data Mining,
  Kluwer Academic Publishers, Norwell, MA, USA, 1998.

\bibitem{koller_toward_1995}
D.~Koller, M.~Sahami, Toward optimal feature selection, in: In 13th
  {International} {Conference} on {Machine} {Learning}, 1995, pp. 284--292.

\bibitem{wu_massive-scale_2014}
Y.~Wu, S.~C.~H. Hoi, T.~Mei, Massive-scale {Online} {Feature} {Selection} for
  {Sparse} {Ultra}-high {Dimensional} {Data}, arXiv:1409.7794 [cs]ArXiv:
  1409.7794.

\bibitem{HuPYL10-ieee-tsmc}
Q.~Hu, W.~Pedrycz, D.~Yu, J.~Lang, Selecting discrete and continuous features
  based on neighborhood decision error minimization, {IEEE} Transactions on
  Systems, Man, and Cybernetics, Part {B} 40~(1) (2010) 137--150.
\newblock \href {http://dx.doi.org/10.1109/TSMCB.2009.2024166}
  {\path{doi:10.1109/TSMCB.2009.2024166}}.

\bibitem{XingJK01-icml}
E.~P. Xing, M.~I. Jordan, R.~M. Karp, Feature selection for high-dimensional
  genomic microarray data, in: Proceedings of the Eighteenth International
  Conference on Machine Learning {(ICML} 2001), Williams College, Williamstown,
  MA, USA, June 28 - July 1, 2001, 2001, pp. 601--608.

\bibitem{baur_feature_2016}
B.~Baur, S.~Bozdag,
  \href{http://journals.plos.org/plosone/article?id=10.1371/journal.pone.0148977}{A
  {Feature} {Selection} {Algorithm} to {Compute} {Gene} {Centric} {Methylation}
  from {Probe} {Level} {Methylation} {Data}}, PLOS ONE 11~(2) (2016) e0148977.
\newblock \href {http://dx.doi.org/10.1371/journal.pone.0148977}
  {\path{doi:10.1371/journal.pone.0148977}}.
\newline\urlprefix\url{http://journals.plos.org/plosone/article?id=10.1371/journal.pone.0148977}

\bibitem{MichaelElad08-SIAM}
M.~Aharon, M.~Elad, Sparse and redundant modeling of image content using an
  image-signature-dictionary, SIAM Journal on Imaging Sciences 1~(3) (2008)
  228--247.
\newblock \href {http://dx.doi.org/10.1137/07070156X}
  {\path{doi:10.1137/07070156X}}.

\bibitem{Forman2003-jmlr}
G.~Forman, An extensive empirical study of feature selection metrics for text
  classification, J. Mach. Learn. Res. 3 (2003) 1289--1305.

\bibitem{hava_supervised_2013}
O.~Háva, M.~Skrbek, P.~Kordík, Supervised two-step feature extraction for
  structured representation of text data, Simulation Modelling Practice and
  Theory 33 (2013) 132--143.
\newblock \href {http://dx.doi.org/10.1016/j.simpat.2012.11.003}
  {\path{doi:10.1016/j.simpat.2012.11.003}}.

\bibitem{liu_information-theoretic_2015}
H.~Liu, Z.~Wu, X.~Zhang, D.~F. Hsu, An information-theoretic feature selection
  method based on estimation of {Markov} blanket, in: 2015 {IEEE} 14th
  {International} {Conference} on {Cognitive} {Informatics} {Cognitive}
  {Computing} ({ICCI}*{CC}), 2015, pp. 327--332.
\newblock \href {http://dx.doi.org/10.1109/ICCI-CC.2015.7259406}
  {\path{doi:10.1109/ICCI-CC.2015.7259406}}.

\bibitem{Liu04-jmlr}
L.~Yu, H.~Liu, Efficient feature selection via analysis of relevance and
  redundancy, Journal of Machine Learning Research 5 (2004) 1205--1224.

\bibitem{vergara_review_2013}
J.~R. Vergara, P.~A. Estévez, A review of feature selection methods based on
  mutual information, Neural Computing and Applications 24~(1) (2013) 175--186.
\newblock \href {http://dx.doi.org/10.1007/s00521-013-1368-0}
  {\path{doi:10.1007/s00521-013-1368-0}}.

\bibitem{Koller_probabilistic_2009}
D.~Koller, N.~Freidman, Probabilistic {Graphical} {Models}: {Principles} and
  {Techniques}, 1st Edition, The MIT Press, Cambridge, MA, 2009.

\bibitem{zhou2005streaming_alphainvesting}
J.~Zhou, D.~Foster, R.~Stine, L.~Ungar, Streaming feature selection using
  alpha-investing, in: Proceedings of the eleventh ACM SIGKDD international
  conference on Knowledge discovery in data mining, ACM, 2005, pp. 384--393.

\bibitem{liu1995chi2}
H.~Liu, R.~Setiono, Chi2: feature selection and discretization of numeric
  attributes, in: Tools with Artificial Intelligence, 1995. Proceedings.,
  Seventh International Conference on, 1995, pp. 388--391.
\newblock \href {http://dx.doi.org/10.1109/TAI.1995.479783}
  {\path{doi:10.1109/TAI.1995.479783}}.

\bibitem{uciLichman2013}
M.~Lichman, \href{http://archive.ics.uci.edu/ml}{{UCI} machine learning
  repository} (2013).
\newline\urlprefix\url{http://archive.ics.uci.edu/ml}

\bibitem{guyon2004result}
I.~Guyon, S.~Gunn, A.~Ben-Hur, G.~Dror, Result analysis of the nips 2003
  feature selection challenge, in: Advances in Neural Information Processing
  Systems, 2004, pp. 545--552.

\bibitem{fayyad_multi-interval_1993}
U.~M. Fayyad, K.~B. Irani, Multi-{Interval} {Discretization} of
  {Continuous}-{Valued} {Attributes} for {Classification} {Learning}, in:
  Proceedings of the {International} {Joint} {Conference} on {Uncertainty} in
  {AI}, 1993, pp. 1022--1027.

\bibitem{cortes_support-vector_1995}
C.~Cortes, V.~Vapnik, Support-{Vector} {Networks}, Machine Learning 20~(3)
  (1995) 273--297.
\newblock \href {http://dx.doi.org/10.1023/A:1022627411411}
  {\path{doi:10.1023/A:1022627411411}}.

\bibitem{Hastie_elements_2011}
T.~J. Hastie, R.~Tibshirani, J.~H. Friedman, The {Elements} of {Statistical}
  {Learning}: {Data} {Mining}, {Inference}, and {Prediction}, 2nd Edition,
  Springer, 2011.

\bibitem{svmlib}
C.-C. Chang, C.-J. Lin, {LIBSVM}: A library for support vector machines, ACM
  Transactions on Intelligent Systems and Technology 2 (2011) 27:1--27:27,
  software available at \url{http://www.csie.ntu.edu.tw/~cjlin/libsvm}.

\end{thebibliography}
\end{document}